\documentclass[conference]{IEEEtran}
\IEEEoverridecommandlockouts
\usepackage[T1]{fontenc}
\usepackage{cite}
\usepackage{amsmath,amssymb,amsfonts}
\usepackage{graphicx}
\usepackage{capt-of}
\usepackage{xcolor}
\usepackage{hyperref}
\usepackage{booktabs}
\usepackage{tikz}

\usetikzlibrary{arrows.meta}
\graphicspath{{figures/}}

\newcommand{\ownds}{UCSD}
\newcommand{\xo}{x^{\mathrm{obs}}}

\begin{document}

\title{Observation-Referenced Tissue Deformation Forecasting with Tool-Driven and Residual Dynamics}
\author{\IEEEauthorblockN{Tom Olesch$^{1}$\thanks{Preprint. This work has been
submitted to the IEEE for possible publication. Copyright may be transferred
without notice, after which this version may no longer be accessible.}, Nan Xiao$^{1}$, Xiongye Xiao$^{1}$,
Jonathan C.~DeLong$^{2}$ and Fei Liu$^{1}$}
\IEEEauthorblockA{$^{1}$Department of Electrical Engineering and Computer Science,
University of Tennessee, Knoxville\\
$^{2}$Department of Surgery, University of Tennessee Medical Center, Knoxville\\
tolesch@vols.utk.edu, nxiao4@vols.utk.edu, xxiao9@utk.edu, fliu33@utk.edu}}
\IEEEaftertitletext{%
\begin{minipage}{\textwidth}
\centering
\includegraphics[width=.98\textwidth]{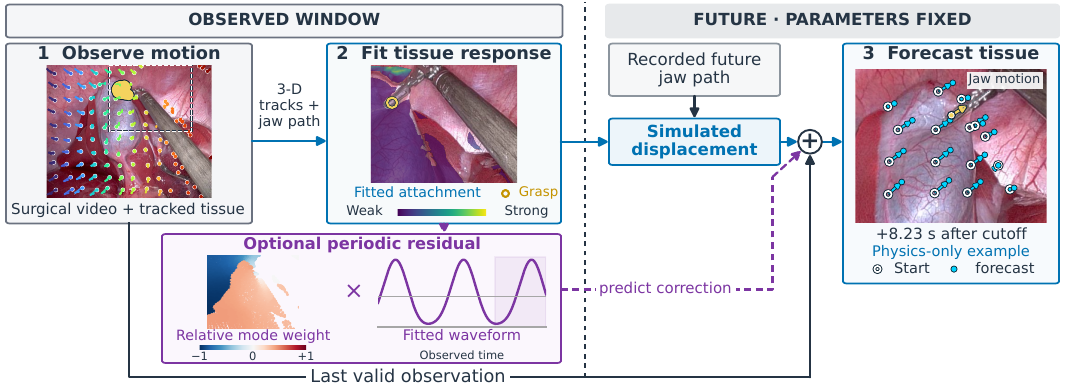}
\captionof{figure}{Observation-referenced tissue forecasting on SM-P3, a calibrated stereo clip from StereoMIS (corpus in Sec.~\ref{sec:experiments}). Middle: fitted dimensionless attachment weights (0--0.3) interpolated over supported tissue, excluding tools and gaps. The ring marks the grasp. Right: true-scale cyan arrows connect white starting rings to physics forecasts over the cutoff image. Yellow shows recorded jaw displacement. The optional periodic residual combines fixed spatial patterns with fitted temporal oscillations. One representative mode and temporal component are illustrated. Spatial weights are normalized to maximum absolute value 1: positive follows the oscillation, negative inverts it, and zero contributes nothing. The dashed box marks the shared crop. Tracks are reconstructed offline, and forecast geometry uses only prefix tissue observations.}
\label{fig:inputs}
\end{minipage}\par\medskip
}
\maketitle

\begin{abstract}
Forecasting tissue deformation from future instrument motion could support predictive planning and increasingly autonomous surgical systems. We study this problem using a scene-specific differentiable surface model fitted to an observed tool-tissue interaction window. Rather than forecasting absolute simulated positions, we introduce an \textit{observation-referenced} formulation that anchors each tissue point to its last valid observation and adds only the deformation induced by the known, preplanned future tool trajectory. This reduces sensitivity to reconstruction, initialization, and persistent simulation offsets while preserving the predicted deformation response. The physical model uses differentiable, mesh-free position-based dynamics with scene-specific rigidity, attachment, region scale, and grasp coupling. We further introduce an optional \textit{observation-referenced residual model} combining graph-spectral spatial modes with periodic harmonic dynamics for structured motion not explained by the tool input. We evaluate the framework retrospectively on 12 surgical manipulation clips from three datasets. Observation referencing provides the largest gain, with a median within-clip error ratio of $0.598$ relative to the corresponding absolute simulation forecast. The optional periodic residual further improves forecasting when the remaining motion exhibits periodic structure. These results support observation-referenced deformation prediction as a practical building block for future surgical world models, digital twins, and generative simulators that must predict how tissue evolves under instrument actions.
\end{abstract}

\begin{figure*}[t]
    \centering
    \resizebox{.98\textwidth}{!}{\begingroup
\input{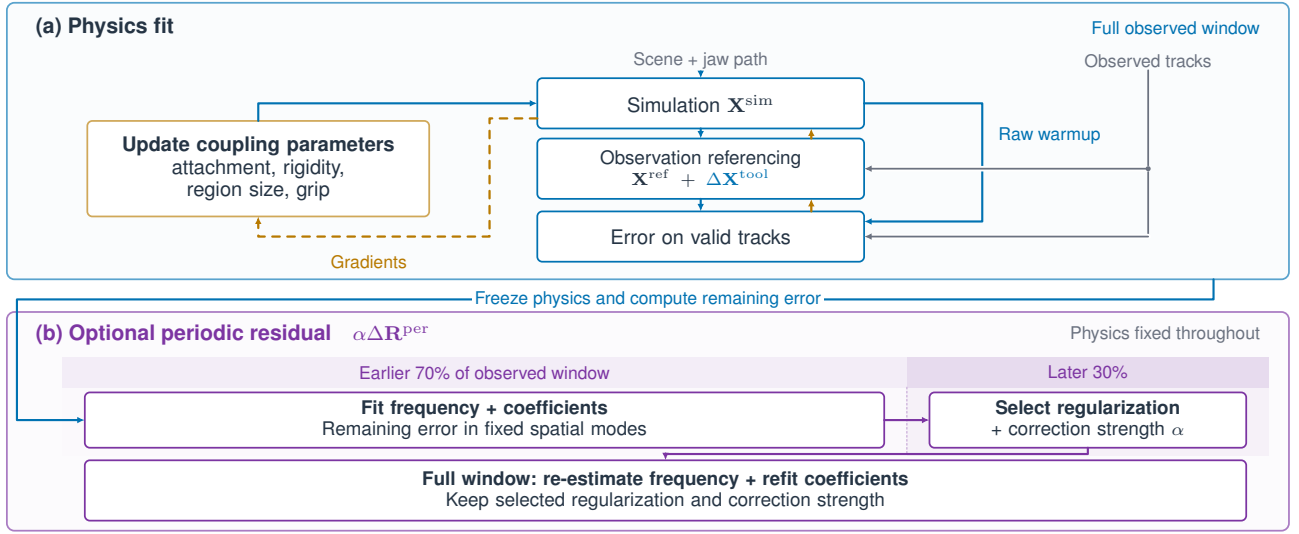}
\begin{tikzpicture}[
  x=1bp,y=-1bp,font=\sffamily\fontsize{8}{9}\selectfont,text=wfDark,
  line cap=round,line join=round,>={Latex[length=3bp,width=2.7bp]},
  flow/.style={->,draw=wfBlue,line width=.85bp},
  grad/.style={->,draw=wfRed,line width=.9bp,dashed},
  observation/.style={->,draw=wfGray,line width=.7bp,
    preaction={draw=white,line width=2.2bp}},
  optional/.style={->,draw=wfPurple,line width=.85bp},
  card/.style={draw=wfBlue!65,fill=wfBlue!2,rounded corners=3bp,line width=.7bp,inner sep=0pt},
  wfbox/.style={draw=wfBlue,fill=white,rounded corners=2bp,align=center,inner sep=2bp,outer sep=0pt,line width=.7bp},
  heading/.style={font=\sffamily\bfseries\fontsize{8.5}{10}\selectfont},
  small/.style={font=\sffamily\fontsize{7}{8}\selectfont,text=wfGray}
]
\path[use as bounding box] (0,0) rectangle (510,210);

\node[card,minimum width=510bp,minimum height=110bp] (physics) at (255,55) {};
\node[heading,anchor=west] at (8,10) {(a) Physics fit};
\node[small,anchor=east,text=wfBlue] at (502,10) {Full observed window};
\node[wfbox,draw=wfRed!65,minimum width=136bp,minimum height=38bp,text width=128bp]
  (update) at (100,66) {\textbf{Update coupling parameters}\\
    attachment, rigidity,\\region size, grip};
\node[wfbox,minimum width=130bp,minimum height=20bp,text width=124bp]
  (simulate) at (276,40) {Simulation $\mathbf{X}^{\mathrm{sim}}$};
\node[wfbox,minimum width=130bp,minimum height=24bp,text width=124bp,
      font=\sffamily\fontsize{7.5}{8.5}\selectfont]
  (reference) at (276,66) {Observation referencing\\
    $\mathbf{X}^{\mathrm{ref}}+\textcolor{wfBlue}{\Delta\mathbf{X}^{\mathrm{tool}}}$};
\node[wfbox,minimum width=130bp,minimum height=20bp,text width=124bp]
  (error) at (276,93) {Error on valid tracks};
\node[small,inner sep=1bp] (scene) at (276,23) {Scene + jaw path};
\node[small,inner sep=1bp] (tracks) at (454,23) {Observed tracks};
\draw[flow] (scene.south) -- (simulate.north);
\draw[flow] (update.north) -- (100,40) -- (simulate.west);
\draw[flow] (simulate.south) -- (reference.north);
\draw[flow] (reference.south) -- (error.north);
\draw[flow] (simulate.east) -- (388,40) -- (388,87) -- (341,87);
\node[small,anchor=north west,align=left,text=wfBlue] at (391,46) {Raw warmup};
\draw[wfGray,line width=.7bp] (tracks.south) -- (454,93);
\draw[observation] (454,66) -- (reference.east);
\draw[observation] (454,93) -- (error.east);
\fill[wfGray] (454,66) circle[radius=1bp];
\draw[grad] ([xshift=44bp]error.north) -- ([xshift=44bp]reference.south);
\draw[grad] ([xshift=44bp]reference.north) -- ([xshift=44bp]simulate.south);
\draw[grad] (211,46) -- (192,46) -- (192,93) -- (100,93) -- (update.south);
\node[small,text=wfRed] at (144,103) {Gradients};

\node[card,draw=wfPurple!65,fill=wfPurple!2,minimum width=510bp,minimum height=87bp]
  (residual) at (255,166.5) {};
\node[heading,anchor=west,text=wfPurple] at (8,132)
  {(b) Optional periodic residual\quad $\alpha\Delta\mathbf{R}^{\mathrm{per}}$};
\node[small,anchor=east] at (502,132) {Physics fixed throughout};
\fill[wfPurple!3] (22,141) rectangle (358,180);
\fill[wfPurple!7] (358,141) rectangle (502,180);
\fill[wfPurple!9] (22,141) rectangle (358,153);
\fill[wfPurple!18] (358,141) rectangle (502,153);
\node[small,text=wfPurple] (earlyLabel) at (190,147) {Earlier 70\% of observed window};
\node[small,text=wfPurple] (lateLabel) at (430,147) {Later 30\%};
\draw[wfPurple!35,densely dotted,line width=.6bp] (358,153) -- (358,180);
\node[wfbox,draw=wfPurple,minimum width=318bp,minimum height=22bp,text width=312bp,
      font=\sffamily\fontsize{7.5}{8.5}\selectfont]
  (early) at (190,166) {\textbf{Fit frequency + coefficients}\\
    Remaining error in fixed spatial modes};
\node[wfbox,draw=wfPurple,minimum width=126bp,minimum height=22bp,text width=120bp,
      font=\sffamily\fontsize{7.5}{8.5}\selectfont]
  (late) at (430,166) {\textbf{Select regularization}\\+ correction strength $\alpha$};
\node[wfbox,draw=wfPurple,minimum width=462bp,minimum height=24bp,text width=456bp,
      font=\sffamily\fontsize{7.5}{8.5}\selectfont]
  (refit) at (262,194) {\textbf{Full window: re-estimate frequency + refit coefficients}\\
    Keep selected regularization and correction strength};
\draw[flow] (480,110) -- (480,118) -- (4,118) -- (4,166) -- (early.west);
\node[small,text=wfBlue,fill=white,inner sep=1bp] at (255,118)
  {Freeze physics and compute remaining error};
\draw[optional] (early.east) -- (late.west);
\draw[optional] (late.south) -- (430,179.5) -- (262,179.5) -- (refit.north);

\foreach \wfNode in {physics,residual,update,simulate,reference,error,earlyLabel,lateLabel,early,late,refit}{
  \pgfextractx{\dimen0}{\pgfpointanchor{\wfNode}{center}}
  \pgfextracty{\dimen2}{\pgfpointanchor{\wfNode}{center}}
  \typeout{WFALIGN-\wfNode-center=\the\dimen0,\the\dimen2}
  \pgfextractx{\dimen0}{\pgfpointanchor{\wfNode}{west}}
  \pgfextractx{\dimen2}{\pgfpointanchor{\wfNode}{east}}
  \typeout{WFALIGN-\wfNode-span=\the\dimen0,\the\dimen2}
}
\end{tikzpicture}
\endgroup}
    \captionof{figure}{Observation-referenced tissue forecasting on SM-P3, a calibrated
    stereo clip from StereoMIS (corpus in Sec.~\ref{sec:experiments}). Middle: fitted
    attachment weights interpolated over supported tissue, with the ring marking the
    grasp. Right: cyan arrows connect last observations (white rings) to physics
    forecasts, yellow shows recorded jaw displacement. The optional periodic residual
    combines fixed spatial modes with fitted temporal oscillations, one of each shown.
    Tracks are reconstructed offline, and forecast geometry uses only prefix
    observations.}
    \label{fig:overview}
    \vspace{-0.5cm}
\end{figure*}

\section{Introduction}

Forecasting soft-tissue deformation under surgical manipulation is important for autonomous robotic assistance, motion planning, and predictive visualization \cite{Attanasio_2021, Schmidgall_2025}. Instrument motion can be planned over a short horizon, while occlusion or limited visibility may interrupt tissue observations. This motivates models that infer tissue response from an observed interaction and forecast deformation under a known, preplanned instrument trajectory.

Physics-based simulation explicitly relates tool motion to tissue response, and differentiable simulation allows scene-specific mechanical and coupling parameters to be identified from observed motion \cite{Michał_2026_RAL,liu2021pbd}. Absolute simulated positions are sensitive to reconstruction error, initialization error, and sim-to-real mismatch. The simulated surface can remain spatially offset even when its predicted deformation trend is informative. Other structured motion, including periodic physiological motion, may also remain unmodeled.

We therefore use simulation to predict \textit{deformation relative to the latest observed tissue configuration}. Each point is anchored to its last valid observation, and a scene-specific, differentiable mesh-free surface model contributes displacement under the known, preplanned future tool motion. This \textbf{observation-referenced} formulation removes constant simulation offsets while preserving the predicted deformation.

An \textbf{optional observation-referenced residual model} captures structured motion remaining after physics fitting. It combines graph-spectral spatial modes with periodic harmonic dynamics, motivated by repetitive physiological motion. Referencing this correction also preserves each point's observed forecast origin. Our contributions are:
\begin{itemize}
    \item An \textbf{observation-referenced formulation} for tool-conditioned tissue deformation forecasting.
    \item Scene-specific differentiable identification using only the observed tool-tissue interaction.
    \item An optional observation-referenced periodic residual extension and a controlled evaluation separating fitting, referencing, and residual modeling.
\end{itemize}

Such models could support synthetic data generation, policy evaluation, and planning in future \textit{surgical world models, digital twins, and generative simulators} that predict tissue response to instrument actions, such as recent work from NVIDIA's Cosmos-H-Surgical Simulator~\cite{he2026cosmoshsurgicallearningsurgicalrobot, Saurabh_2025_MICCAI,he2026cosmoshsurgicallearningsurgicalrobot}.

\section{Related Work}

\subsection{Video-derived surgical simulation}
Video-derived reconstruction has enabled increasingly detailed models of dynamic surgical scenes. Gaussian-splatting methods reconstruct and render endoscopic tissue motion~\cite{endogaussian,deform3dgs}, while physics-embedded approaches incorporate tissue mechanics and, in some cases, estimate material parameters~\cite{simendogs,endogsim,surgisim}. Position-based tissue registration couples surgical perception with
deformable simulation~\cite{liu2021pbd}, evaluated by registration fit to observed frames rather than forecasting unseen motion.
We evaluate held-out tissue forecasts under preplanned instrument motion, separating the contributions of parameter fitting and observation referencing.

\subsection{Differentiable and Position-Based Simulation}
Position-based methods project positions directly onto constraint manifolds
for unconditionally stable updates~\cite{muller2007pbd,xpbd}, extended to
stable Neo-Hookean elasticity through a compliant constraint
formulation~\cite{smith2018snh,macklin2021}. constraints and stiffness are
specified in advance rather than identified from observation. Shape
matching~\cite{shapematch,fastlsm} and embedded
deformation~\cite{embedded} similarly propagate coherent surface motion, but
from prescribed handle motion rather than data.

Differentiable projective dynamics~\cite{diffpd} and simulation-based
identification~\cite{gradsim} instead optimize simulator parameters from
observations, and learned mesh simulators such as
MeshGraphNets~\cite{meshgraphnets} predict deformation via graph message
passing trained on many rollouts. Both target synthetic or richly
instrumented data, not one short surgical window with no future
supervision. We combine regional shape matching with differentiable
identification to fit a small, physically-grounded parameter set from
exactly that regime.

\section{Methodology}

\subsection{Problem Formulation and Forecast Decomposition}

We forecast tissue motion from an observed tool-tissue interaction and a known future instrument path. Given tissue observations $\mathbf{X}^{\mathrm{obs}}_{0:\tau}$ up to cutoff $\tau$ and instrument trajectory $\mathbf{U}_{0:T}$ over the full sequence, we predict future tissue trajectories $\hat{\mathbf{X}}_{\tau+1:T}$. Tissue observations after $\tau$ are withheld for evaluation.

We decompose the forecast as
\begin{equation}
\hat{\mathbf{X}}_{\tau+1:T}
=
\mathbf{X}^{\mathrm{ref}}
+
\Delta\mathbf{X}^{\mathrm{tool}}
+
\alpha\Delta\mathbf{R}^{\mathrm{per}},
\label{eq:forecast_decomposition}
\end{equation}
where $\mathbf{X}^{\mathrm{ref}}$ contains each point's last valid observed position, $\Delta\mathbf{X}^{\mathrm{tool}}$ is displacement predicted by the physical model under known, preplanned instrument motion, and $\alpha\Delta\mathbf{R}^{\mathrm{per}}$ models periodic structure in the remaining fitting error, motivated by physiological motion~\cite{mcclelland}. The residual is also referenced to each point's last valid observation, with correction strength $\alpha\in[0,1]$.

Thus, the physical model captures tool-driven tissue deformation. Periodic physiological motion can introduce additional deformation independent of tool input.
We fit physics first, then residual dynamics with physics fixed (shown in Fig.~\ref{fig:overview}). All parameters use the observed interval and remain fixed during forecasting.

\begin{figure*}[!ht]
\centering
    \includegraphics[width=.98\textwidth]{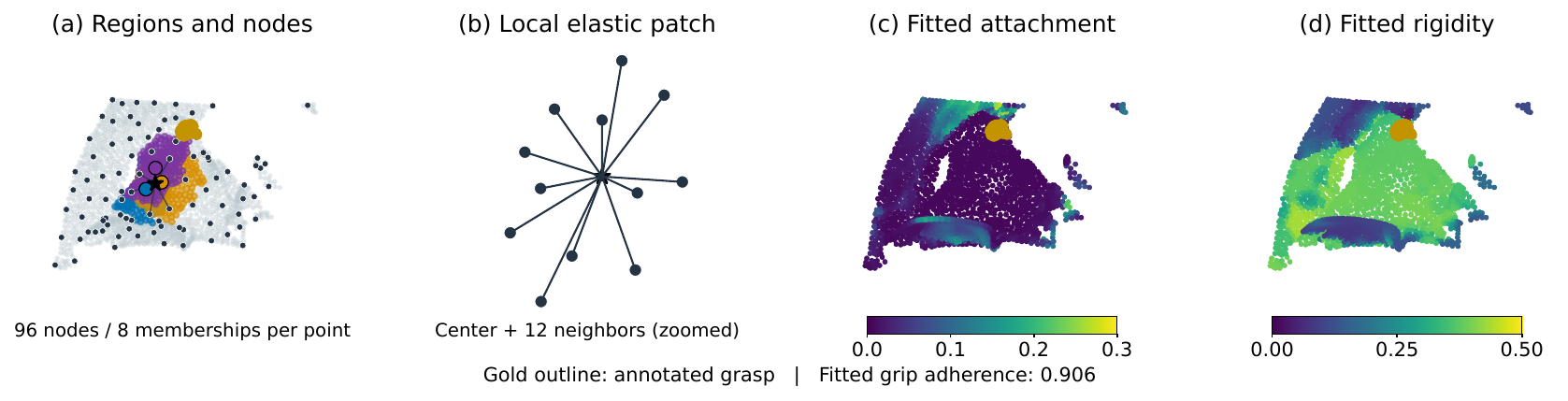}
    \caption{Simulation structure and identified coupling on the SM-P3 example clip (Sec.~\ref{sec:experiments}) in a camera-oriented 3-D view. (a) All 96 scaffold nodes with 3 example regions and the 8 point-to-node memberships of the starred point.
    (b) Its local elastic patch connects the center to 12 neighbors.
    (c,d) Colors show fitted attachment and membership-weighted regional rigidity,
    starting from uniform values of 0.05 and 0.15, respectively.
    Gold outlines mark the annotated grasp, whose shared adherence is fitted.
    All coupling values are dimensionless and fitted on the observed prefix.}
    \label{fig:structure}
    \vspace{-0.5cm}
\end{figure*}
\subsection{Tool-Driven Deformation Dynamics}
The tool-driven deformation term $\Delta\mathbf{X}^{\mathrm{tool}}$ in Eq.~\eqref{eq:forecast_decomposition} is predicted using a scene-specific differentiable surface model. Reconstructed 3-D tissue tracks define the deformable surface, while tracked instrument motion provides the driving input. With parameters $\boldsymbol{\theta}$, the model generates
\begin{equation}
\mathbf{X}^{\mathrm{sim}}_{0:T}
=
\mathcal{S}_{\boldsymbol{\theta}}
\left(
\mathbf{X}_0,\mathbf{U}_{0:T}
\right).
\label{eq:simulator}
\end{equation}
Here, $\mathbf{X}_0$ denotes the stored initial surface estimate used to initialize the simulation, while the fixed reference cloud $\mathbf{X}$ defines the reference geometry used by the deformation model. The two may differ when the retained observation sequence does not begin at the original reference state.

\subsubsection{Surface Representation and Tool Coupling}

The reconstructed tissue surface is represented by a fixed reference cloud $\mathbf{X}$ with $N$ points, local elastic neighborhoods, and an overlapping regional scaffold of $M$ nodes. Each point belongs to its neighboring scaffold nodes with membership weight $w_{im}=\exp\left(-\|\mathbf{X}_i-\mathbf{X}_m\|^2/(2r_m^2)\right)$, where $r_m$ controls distance weighting within the fixed membership of region $m$. Local neighborhoods capture short-range deformation, while overlapping regional shape matching coordinates motion across larger spatial regions.

Tracked jaw translation provides the tool input. While the simulator supports full 6-D tool motion, reliable rotational estimates require robust 6-D pose tracking that is not consistently available from the current single-view video-based tracking pipeline. We therefore use translational motion to drive the grasp constraints and keep the jaw orientation fixed during each episode. The annotated grasp set $\mathcal{G}$ and each grasp point's initial offset relative to the jaw are fixed throughout the episode. For $i\in\mathcal{G}$, the grasp target is defined as $\mathbf{g}_i(t)=\mathbf{x}^{\mathrm{obs}}_i(0)+\mathbf{u}(t)-\mathbf{u}(0)$, where $\mathbf{u}(t)$
denotes the tracked jaw translation.

A compliant grasp constraint couples points in $\mathcal{G}$ to the moving tool target, while per-point attachment constraints represent spatially varying anchoring to the surrounding tissue. Specifically, their positional corrections are
\begin{equation}
\Delta\mathbf{x}^{\mathrm{att}}_i
=
a_i(\mathbf{X}_i-\mathbf{x}_i),
~
\Delta\mathbf{x}^{\mathrm{grasp}}_i
=
g(\mathbf{g}_i(t)-\mathbf{x}_i),~ i\in\mathcal{G},
\label{eq:attachment_grasp}
\end{equation}
where $a_i$ is the fitted attachment strength and $g$ is the shared grasp adherence. Thus, attachment regulates how strongly each surface point remains coupled to its reference configuration, while grasp adherence controls how closely the grasped tissue follows the instrument motion.

\subsubsection{Deformation Constraints}

Local deformation is measured relative to each point's reference neighborhood. For reference offsets $\mathbf{r}_{ij}=\mathbf{X}_j-\mathbf{X}_i$, we use fixed distance weights $a_{ij}=\exp\left(-\|\mathbf{r}_{ij}\|^2/\sigma^2\right)$, where $\sigma$ is the median reference-neighbor distance, and define $\mathbf{B}_i=\sum_j a_{ij}\mathbf{r}_{ij}\mathbf{r}_{ij}^{T}$. The local deformation gradient is
\begin{equation}
\mathbf{F}_i(\mathbf{x})
=
\left[
\sum_j
a_{ij}
(\mathbf{x}_j-\mathbf{x}_i)
\mathbf{r}_{ij}^{T}
\right]
\left(
\mathbf{B}_i+\epsilon_B\mathbf{I}
\right)^{-1},
\label{eq:deformation_gradient}
\end{equation}
where $\epsilon_B=10^{-6}$ regularizes nearly planar neighborhoods. This perturbs the reference-state gradient from identity, so the elastic energy acts as an effective surface regularizer.

We impose stable Neo-Hookean elasticity~\cite{smith2018snh} using the first-order constraint formulation of Macklin and M\"uller~\cite{macklin2021, Samuel_2026}. Specifically, $C_h(\mathbf{F})=\det(\mathbf{F})-\gamma$ and $C_d(\mathbf{F})=\sqrt{\|\mathbf{F}\|_F^2+\epsilon_d}$, with $\gamma=1+\mu/\lambda$, $\mu=\lambda=1$, and $\epsilon_d=10^{-6}$. Their derivatives are $\partial C_h/\partial\mathbf{F}=\operatorname{cof}(\mathbf{F})$ and $\partial C_d/\partial\mathbf{F}=\mathbf{F}/C_d(\mathbf{F})$. These local elastic constraints regulate short-range deformation, while regional shape matching maintains coherent motion across overlapping neighborhoods by aligning each reference region to the current simulated surface and blending the resulting targets through the membership weights.

The surface is advanced quasi-statically between video frames. At each frame, XPBD multipliers are reset and 20 projection iterations are applied in the order of attachment, regional shape matching, local elasticity, and grasp. Jacobi corrections are relaxed by $0.2$ and capped at $0.005$ solver units per point per projection, with compliance evaluated using the fixed numerical scale $h=1/240$. The recorded frame cadence is retained for each sequence. Because the reconstructed surface is only approximately stress-free under the regularized local model and finite projection updates, small rest-state drift can remain, as examined in Sec.~\ref{sec:validation}.

\subsubsection{Scene-Specific Parameter Identification}

We fit effective coupling parameters to the observed interaction. The set $\boldsymbol{\theta}=\{\boldsymbol{\rho},\mathbf{r},\mathbf{a},g\}$ comprises regional rigidity $\boldsymbol{\rho}\in\mathbb{R}^{M}$, region scales $\mathbf{r}\in\mathbb{R}^{M}$, per-point attachment $\mathbf{a}\in\mathbb{R}^{N}$, and shared grasp adherence $g$. Logistic and logarithmic parameterizations constrain rigidity to $(0,0.5)$, attachment to $(0,1)$, grasp adherence to $(0.05,1)$, and scales to positive values.

To make identification consistent with observation-referenced forecasting, we fit simulated changes relative to each point's last valid observation. Let $v_i(t)\in\{0,1\}$ indicate observation validity. For points observed in the prefix, $t_i^\star=\max\{t\leq\tau:v_i(t)=1\}$ is the last valid observation index. This point-specific reference accommodates missing or occluded observations whose final valid frame may precede $\tau$. We define the referenced prediction
\begin{equation}
\mathbf{q}_i(t;\boldsymbol{\theta})
=
\mathbf{x}^{\mathrm{obs}}_i(t_i^\star)
+
\mathbf{x}^{\mathrm{sim}}_i(t;\boldsymbol{\theta},\mathbf{U})
-
\mathbf{x}^{\mathrm{sim}}_i(t_i^\star;\boldsymbol{\theta},\mathbf{U}).
\label{eq:referenced_simulation}
\end{equation}
Here, $\mathbf{x}^{\mathrm{obs}}_i(t_i^\star)$ anchors the prediction to the last valid observed position, while the difference between the two simulated states provides the model-predicted deformation relative to that reference. We minimize the masked trajectory error over the observed interval,
\begin{equation}
\mathcal{L}_{\mathrm{fit}}(\boldsymbol{\theta})
=
\frac{
\sum_{t=0}^{\tau}\sum_i
v_i(t)
\left\|
\mathbf{q}_i(t;\boldsymbol{\theta})
-
\mathbf{x}^{\mathrm{obs}}_i(t)
\right\|_2^2
}{
\sum_{t=0}^{\tau}\sum_i v_i(t)
},
\label{eq:fit_loss}
\end{equation}
where the validity mask $v_i(t)$ ensures that only available tissue observations contribute to the objective.

We first perform a 30-epoch masked raw-trajectory warmup, followed by 60 epochs using the observation-referenced objective in Eq.~\eqref{eq:fit_loss}. Gradients are propagated through the full simulation rollout, with eight-frame checkpoint segments used for memory efficiency. We optimize with Adam using a learning rate of $0.03$ and gradient-norm clipping at $0.5$, and retain the lowest-loss parameter state as $\boldsymbol{\theta}^{\star}$.

\subsubsection{Tool-Driven Deformation Forecast}

With fitted parameters $\boldsymbol{\theta}^{\star}$ frozen, we extract simulated displacement under the known, preplanned future tool trajectory:
\begin{equation}
\Delta\mathbf{x}^{\mathrm{tool}}_i(t)
=
\mathbf{x}^{\mathrm{sim}}_i
\left(
t;\boldsymbol{\theta}^{\star},\mathbf{U}
\right)
-
\mathbf{x}^{\mathrm{sim}}_i
\left(
t_i^\star;\boldsymbol{\theta}^{\star},\mathbf{U}
\right),
~ t>\tau.
\label{eq:tool_deformation}
\end{equation}
The observed tissue position provides the reference state, and the simulator contributes the predicted change from that state. Together, they give the physics-only forecast in Eq.~\eqref{eq:forecast_decomposition} with $\alpha=0$.

Observation referencing preserves the simulated state. If $\mathbf{e}_i(t)=\mathbf{x}^{\mathrm{sim}}_i(t)-\mathbf{x}^{\mathrm{obs}}_i(t)$ is the absolute simulation error, the referenced forecast error is $\mathbf{e}_i(t)-\mathbf{e}_i(t_i^\star)$. Constant per-point offsets cancel, while displacement errors remain. Zero simulated displacement recovers persistence.

\subsection{Observation-Referenced Residual Forecasting}

The residual model captures spatially coherent periodic motion remaining after physics fitting, supplying the optional term $\alpha\Delta\mathbf{R}^{\mathrm{per}}$ in Eq.~\eqref{eq:forecast_decomposition}. To represent this motion without assigning an independent temporal model to every tissue point, we construct a graph-spectral basis on the reference surface. The basis $\boldsymbol{\Phi}\in\mathbb{R}^{N\times K}$ contains the first $K=32$ modes of the reference cloud's symmetric 12-neighbor combinatorial graph Laplacian, inspired by manifold harmonics~\cite{valletlevy}. Each column describes a fixed spatial pattern whose temporal amplitude is fitted using two harmonics of a shared frequency $\nu$. The residual displacement is
\begin{equation}
\mathbf{r}_i(t)
=
\sum_{k=1}^{K}
\Phi_{ik}
\sum_{h=1}^{2}
\left[
\mathbf{a}_{kh}\cos(2\pi h\nu t)
+
\mathbf{b}_{kh}\sin(2\pi h\nu t)
\right],
\label{eq:periodic_residual}
\end{equation}
where $\Phi_{ik}$ is the contribution of spatial mode $k$ at point $i$, $\nu$ is the fundamental frequency in cycles per frame, $h$ indexes the temporal harmonics, and $\mathbf{a}_{kh},\mathbf{b}_{kh}\in\mathbb{R}^{3}$ are the fitted 3-D harmonic coefficients.

As with the tool-driven deformation, the residual is applied relative to the last valid observation rather than as an absolute displacement. The final forecast is therefore
\begin{equation}
\hat{\mathbf{x}}_i(t)
=
\mathbf{q}_i(t;\boldsymbol{\theta}^{\star})
+
\alpha
\left[
\mathbf{r}_i(t)-\mathbf{r}_i(t_i^\star)
\right],
\qquad t>\tau,
\label{eq:sequential}
\end{equation}
where $\mathbf{r}_i(t_i^\star)$ is the residual predicted at the observation reference and $\alpha\in[0,1]$ controls the correction strength. Subtracting $\mathbf{r}_i(t_i^\star)$ makes the residual correction zero at the reference time, so any residual offset already contained in the observed position is not added again.

The harmonic coefficients are fitted at unit correction strength so that $\mathbf r_i(t)-\mathbf r_i(t_i^\star)$ approximates the remaining observed error $\mathbf{x}^{\mathrm{obs}}_i(t)-\mathbf{q}_i(t;\boldsymbol{\theta}^{\star})$, using masked least squares with a trace-scaled ridge penalty. The model contains 384 harmonic coefficients. The frequency $\nu$, ridge regularization, and correction strength $\alpha$ are selected using only the observed interval, as described in Sec.~\ref{sec:diagnostics}. After selection, frequency and coefficients are refitted on the complete observed prefix. Setting $\alpha=0$ recovers the tool-driven physics forecast alone.

\section{Experiments}
\label{sec:experiments}
\subsection{Corpus, units, and retrospective scope}
We retrospectively evaluate twelve development-selected clips with usable
tracked surfaces and grasp motion: nine monocular CMR Versius clips from seven
Open-H-Embodiment surgeries~\cite{openh} (clips labeled OH-$n$), two calibrated stereo clips from two StereoMIS surgeries~\cite{stereomis} (SM-P2 and SM-P3), and one \ownds{} side-by-side clip with nominal reconstruction scale
(labeled \ownds), from a deidentified recording collected under a UCSD
Human Research Protections Program protocol (IRB \#070449) and annotated
at the University of Tennessee. Source identifiers define ten surgery-level clusters.
Within-clip error ratios allow aggregation across different coordinate scales.
Episode coordinates are stored in units of $0.1$\,m (a factor of $10$ applied
during dataset creation), so calibrated-stereo errors convert to millimeters
by a factor of $100$. The same factor defines the nominal units reported for
monocular and uncalibrated scenes.

We construct tissue trajectories from long-range AllTracker~\cite{alltracker}
image tracks and FoundationStereo~\cite{foundationstereo} depth for stereo recordings
(StereoMIS, \ownds) or Video Depth Anything~\cite{vda} for monocular Open-H clips.
Tool/occluder masks exclude instrument points. Persistent identities propagate
the manually selected grasp set through time. All twelve episodes use
manually marked, tracked instrument points for the jaw trajectory.
The simulator assumes a fixed camera frame, so camera motion during the
\ownds{} pull makes this clip a stress case for that assumption.

Forecast horizons vary across clips, whose cutoffs were fixed near their
midpoints before evaluation. Fitting windows span 0.75--31.50\,s, with forecasts of
0.72--31.47\,s. SM-P2 provides the shortest forecast, while SM-P3 spans
8.23\,s. Durations use the 30 Hz extraction rate for Open-H and \ownds{} and
recording timestamps for StereoMIS (60 Hz for P3 and variable cadence for P2).

The primary evaluation uses a fixed surface subsample selected using
whole-clip visibility. For points that are invalid at the first retained
frame, stored reconstruction estimates initialize the latent surface, while
fitting and scoring use valid observations.

To assess sensitivity to this selection, a three-clip diagnostic samples
2000 points using visibility at frame zero and at least 90\% visibility
within the fitting prefix, including eligible grasp points. Both selections
retain the offline tracking, depth inference, smoothing, and annotation
protocol.

\emph{Implementation} Each scene uses $2000$ surface points with 12-neighbor elastic patches and $M=96$ farthest-point scaffold nodes, each point belonging to its eight nearest nodes, so the fitted parameter set contains $2M+N+1=2193$ values. The complete twelve-clip evaluation uses the tool-driven model alone ($\alpha=0$). The periodic residual extension is evaluated separately on the three manipulation pilots (Sec.~\ref{sec:diagnostics}).

\subsection{Baselines with a common observation cutoff}
All methods use the same tracked points, fitting interval, and future validity
mask. Instrument-conditioned methods receive the same jaw translation.
To separate fitting from referencing, four physics variants cross shared
initial or fitted parameters with raw or observation-referenced forecasts:
\emph{default raw}, \emph{default incremental}, \emph{identified raw}, and
\emph{identified incremental}.

The observation-only baselines extrapolate from the tissue tracks.
\emph{Persistence} holds each point at its last valid observation $\xo_i(t_i^*)$, while
\emph{constant velocity} fits each point's recent valid positions by least
squares in recording time, then extrapolates from its last observation.
Its window is chosen from $\{3,5,10,20,40\}$ frames using a 70/30 temporal
split inside the fitting prefix.

\emph{Tool ridge} predicts each point's response through a linear map from
jaw displacement to tissue displacement, using trace-normalized ridge penalties
$\{10^{-3},10^{-2},10^{-1},1,10\}$ and the same internal temporal split.
After refitting on the full prefix, it predicts
\begin{equation}
 \hat x_i^{\rm ridge}(t)=\xo_i(t_i^*)+
                         A_i^{\mathsf T}[u(t)-u(t_i^*)].
\end{equation}
Here $A_i\in\mathbb R^{3\times3}$ is fitted independently for each point.

\subsection{Metrics and uncertainty}
We score forecasts by their average distance to valid future reconstructed
points. For clip $c$, mean point error is
\begin{align}
 E_c&=\frac{\sum_{t>\tau}\sum_i m_i(t)
                 \|\hat x_i(t)-\xo_i(t)\|_2}
                 {\sum_{t>\tau}\sum_i m_i(t)},\\
 m_i(t)&=v_i(t)\,\mathbf 1[t_i^*\text{ exists}].
\end{align}
All methods share this mask and denominator, with equal weight for valid
point--frame observations. The primary aggregate is the median of
twelve within-clip ratios $E_c^{\rm method}/E_c^{\rm baseline}$. Values below
one favor the numerator.

We report all clips, improvement counts, and 95\% percentile intervals from
10,000 surgery-cluster bootstrap draws (seed 8021), retaining all clips of
each resampled surgery. The ten available clusters give a coarse estimate of
uncertainty within this corpus.
In a post-hoc analysis, we also measure forecast error in pixels after camera projection, using the same cutoffs, masks, and predictions. We additionally evaluate tool ridge fitted directly to projected 2-D tracks, with regularization selected within the observed window.

\subsection{Diagnostics and periodic extensions}
\label{sec:diagnostics}
Three pilot clips designated in advance test sensitivity to initialization and residual
fitting: SM-P2 (short prefix), OH-747
(a failure case identified during development), and SM-P3 (longer calibrated
manipulation).
For each, two additional runs perturb all log/logit initial parameters by
independent Gaussian noise with standard deviation $0.15$ (seeds 1 and 2).
The scaffold topology, data, optimization schedule, and canonical default
baseline remain fixed.

Residual frequency is estimated from observed physics error using the main
rollout and the joint ablation's exact spatial basis. We mean-center the raw
residual's modal coefficients over the observed prefix and select the highest
summed Fourier power among 512 frequencies in $[1/200,1/80]$ cycles per frame.
This inherited search zero-fills invalid residuals, while coefficient fitting
and scoring retain validity masks. The physical frequency band therefore
depends on recording rate.

We select residual settings through a temporal split of the observed prefix.
We fit frequency and coefficients on the first 70\% of frames, then choose ridge strength
$\beta\in\{0,10^{-4},10^{-3},10^{-2},0.1,1,10\}$ and correction strength
$\alpha\in\{0,0.25,0.5,0.75,1\}$ by mean point error on the remaining frames. The penalty adds
$\beta$ times the normal matrix's mean diagonal to each diagonal entry. With
these settings, we refit frequency and coefficients on the complete prefix.
This validation is conditional on physics already fitted to the full observed
prefix, including the residual-validation segment. We recorded the grids and
all new fitting choices before evaluating the sequential future forecasts.
Future tissue coordinates are used only for final scoring.

The \emph{nested sensitivity} also withholds the validation segment from
physics fitting. We refit physics on the first 70\% of each pilot's prefix
with the original 30+60-epoch schedule, fit the residual on this segment,
and select ridge and correction strength on the remaining 30\% with the same
grid. Final forecasts reuse the independent full-prefix physics fit and
refit residual frequency and coefficients on that prefix. Reconstruction
remains fixed and offline.

The \emph{joint-fitting ablation} updates physics and residual coefficients
together. The residual after a 30-epoch physics warmup sets a fixed harmonic
frequency for 60 joint epochs, with coefficient learning rate $0.02$.
We evaluate both the physics component and the full hybrid. Sequential controls
use the same frequency with unrestricted least squares or matched 60-step
coefficient-only Adam. The extended record details all variants and optimizer
settings.

\subsection{Passive periodic-motion case study}
We test how observation-window length affects periodic prediction in a 99.97\,s passive SM-P3 stereo recording using a protocol fixed after exploratory analysis and before evaluation. At each of 3 cutoffs, observation windows of $\{3,6,10,17,27,37\}$\,s precede the same 13\,s forecast, with fixed reference geometry and scoring masks. Primary points have at least 95\% visibility over both the analysis span and the final 3\,s observation window. A sensitivity analysis selects points from the full cloud using only final-window visibility. Each point's mean valid position in this common window defines its reference.

Passive motion combines fixed spatial modes with fitted temporal waveforms. Coefficients in a 32-mode, 12-neighbor graph basis are fitted by least squares over visible vertices. Frequency is estimated within each observation window in the 0.10--0.67\,Hz band after cubic detrending. We then fit 3 harmonic pairs, a constant, and a linear trend. Baselines are persistence and the same modal linear model without harmonics. Fits use recording timestamps, and scoring uses masked mean point error in millimeters.

\section{Results}
\subsection{Numerical validation}
\label{sec:validation}
Constraint derivatives and rollout gradients agree with finite differences in the tested cases. Float32 replays of up to 80 transitions show discrepancies of up to $3.89\times10^{-5}$ solver units, exceeding the initial absolute tolerance of $2\times10^{-5}$. The extended record provides the detailed derivative, replay, and rest-state checks.

\subsection{Complete-corpus forecasting}

\begin{figure}[t]
    \centering\includegraphics[width=\columnwidth]{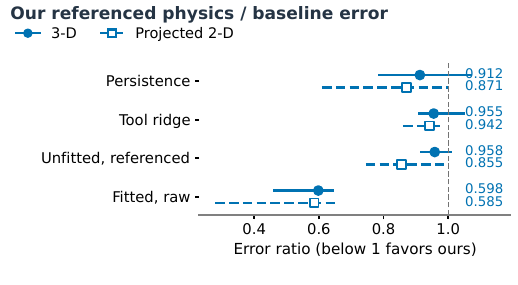}
    \caption{Our fitted, observation-referenced physics relative to four
    baselines across all 12 clips, scored in 3-D and projected image space.
    Markers show median error ratios and bars show 95\% surgery-clustered
    intervals. Ratios below 1 favor our forecast. Both metrics use the same
    frozen forecasts, including the original 3-D-fitted tool ridge.}
    \label{fig:comparison}
    \vspace{-0.5cm}
\end{figure}

\begin{figure*}[!t]
    \centering\includegraphics[width=.94\textwidth]{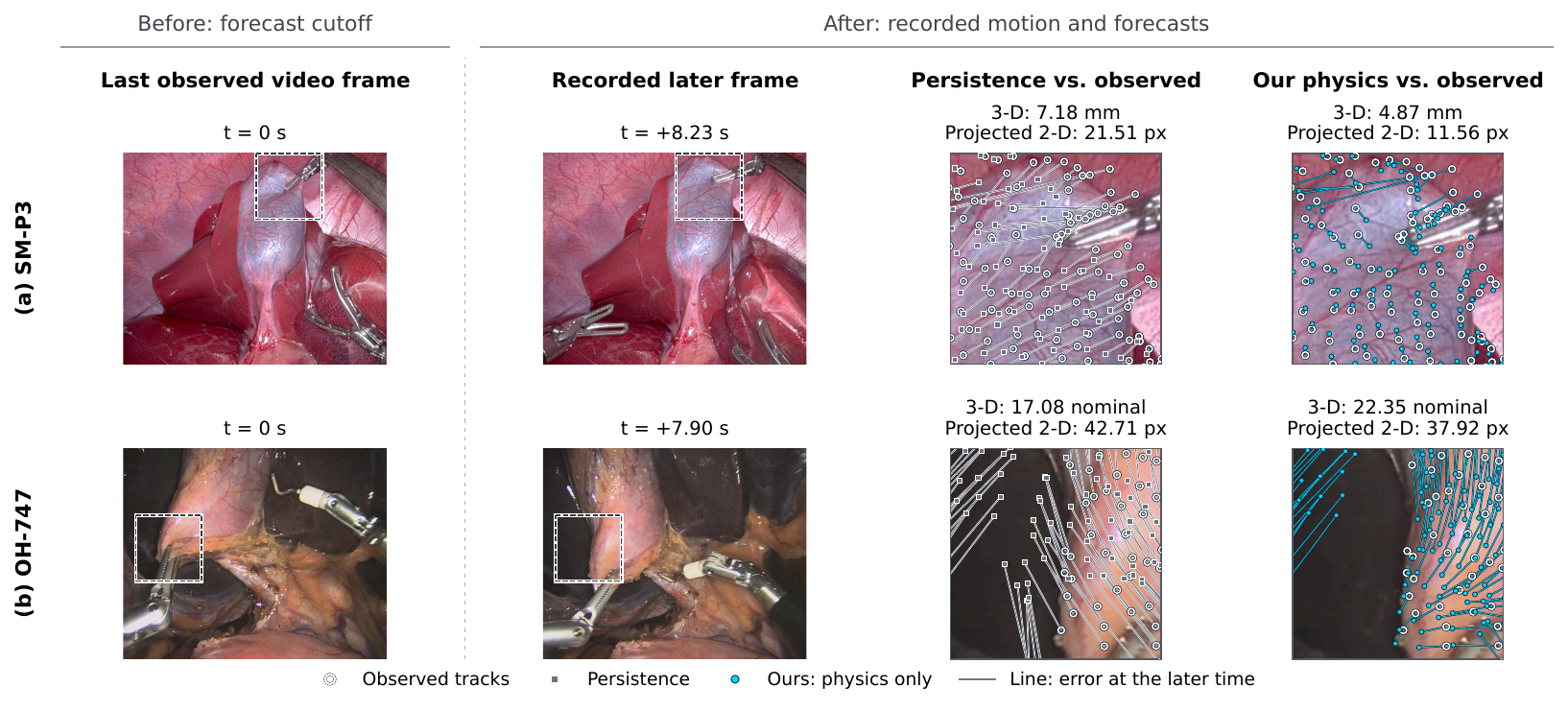}
    \caption{Tissue motion from cutoff to final forecast frame. The first two columns show recorded motion with a fixed, prefix-defined crop. White rings denote reconstructed observations, gray squares persistence, and cyan dots referenced physics. Lines connect final-frame predictions and observations. Labels report final-frame 3-D and image-space errors over all valid points with the same mask, while sampled overlays show local detail.}
    \label{fig:qualitative}
    \vspace{-0.5cm}
\end{figure*}

Simple baselines remain competitive in both 3-D and image space. In 3-D, median error ratios against persistence and tool-driven ridge are respectively 0.912 (95\% interval 0.787--1.070) and 0.955 (0.911--1.048), with 8/12 clips improving in each comparison. Both intervals include equal performance. In image space, referenced physics improves 9/12 clips over persistence (median ratio 0.871, 95\% interval 0.609--1.000). Against ridge fitted in 2-D, the ratio is 0.948 (0.890--1.051), with 8/12 clips improving.

Clip-level results vary, with SM-P2 and SM-P3, the two calibrated stereo clips, showing the largest improvements over persistence. Their error ratios are 0.718 and 0.677, and both also outperform ridge. This observation emerged from the complete-corpus results. For SM-P2, the 0.72\,s forecast limits this finding to a sub-second horizon. In contrast, OH-747 illustrates how image-space and geometric agreement can differ (Fig.~\ref{fig:qualitative}). Referenced physics has higher 3-D error than persistence (14.89 versus 11.91 nominal units), but slightly lower image-space error (24.35 versus 24.96 pixels). The depth coordinate accounts for 97\% of the additional mean squared 3-D error. Independent geometry is needed to separate reconstruction and simulation errors.

\begin{table}[!b]
    \centering
    \begin{tabular}{lrrr}
    \toprule
    Variant & SM-P2 & SM-P3 & OH-747 \\
    \midrule
    Persistence & 1.000 & 1.000 & \underline{1.000} \\
    Physics only (ours) & \underline{0.718} & 0.677 & 1.251 \\
    Joint hybrid & 0.893 & 0.828 & 1.296 \\
    Sequential, unrestricted & 1.333 & 0.725 & 1.338 \\
    Physics + periodic (ours) & 0.731 & \underline{0.668} & 1.189 \\
    Sequential, nested & 0.729 & 0.716 & 1.205 \\
    \bottomrule
    \end{tabular}

    \caption{Periodic extensions on 3 manipulation pilots: held-out 3-D error divided by persistence error (lower is better). Our physics-plus-periodic extension uses conditional selection. Underlines mark column minima. Points, cutoffs, and masks are shared.}
    \label{tab:periodic}\centering\footnotesize
    \vspace{-0.5cm}
\end{table}

\subsection{Training, initialization, and surface selection}
Error ratios to persistence vary little across the three tested initializations: 0.716--0.718 for SM-P2, 0.677--0.681 for SM-P3, and 1.245--1.252 for OH-747.

In the separate surface-selection diagnostic, prefix-selected surfaces yield error ratios of 0.813, 0.689, and 1.233, respectively, with scores also reflecting changes in point sets and discretization. The extended record retains every designated diagnostic run.

\subsection{Periodic motion and residual forecasting}
\emph{Manipulation residuals.}
Conditional residual selection provides modest, clip-dependent gains over physics in 2/3 pilots, versus 1/3 for fully nested selection (Table~\ref{tab:periodic}). The joint hybrid increases error relative to its physics component, and unrestricted sequential fitting increases error relative to main physics. Both selected sequential variants outperform the joint hybrid on all 3 clips, although OH-747 remains worse than persistence. The extended record gives settings and time curves. In image space, conditional residual selection improves on physics in 1/3 pilots, while joint fitting and fully nested selection increase error in all 3.

\begin{figure}[!htb]
    \centering
    \includegraphics[width=\columnwidth]{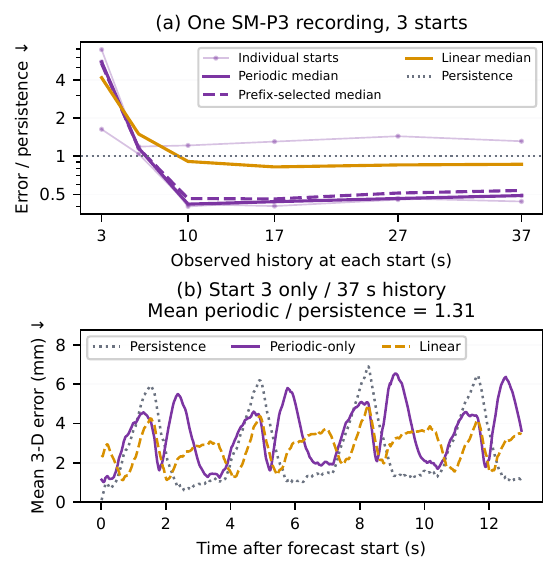}
    \caption{One passive SM-P3 recording, 3 forecast starts. (a) Observation windows vary before each fixed 13\,s forecast. Faint purple curves show individual starts, solid purple/amber show periodic/linear medians, and dashed purple uses prefix-selected points. Ratios below 1 beat persistence. (b) Error at the prespecified last start with 37\,s observed. Here periodic forecasting worsens average error. These passive models are independent of manipulation physics.}
    \label{fig:passive}
    \vspace{-0.5cm}
\end{figure}

\emph{Passive motion.}
Within this recording, extending the observation window from 3\,s to 37\,s reduces the median harmonic-to-persistence error ratio from 5.540 to 0.489 (Fig.~\ref{fig:passive}). At 37\,s, ratios span 0.439--1.312, improving 2/3 cutoffs but worsening the last. Errors vary nonmonotonically with window length. The nonperiodic modal linear and prefix-selected harmonic models yield median ratios of 0.864 and 0.537, respectively, at 37\,s.

The manipulation prefixes contain few fitted cycles: 0.50 cycles of the fitted model on SM-P2, 1.19 on OH-747, and 2.47 on SM-P3. All three frequency estimates lie at the search-band boundaries. Relating these fitted oscillations to respiration would require an independent respiratory signal, which is unavailable for these clips.

\subsection{Compute and reproducibility}
The 90-epoch fits take 2.3--103.8 minutes (median 17.7) and peak at 0.54 GiB GPU memory on an RTX 4070 SUPER (12 GB, PyTorch 2.5.1), including contention but excluding preprocessing. After compilation warmup, simulation runs at about 131 frames/s, excluding reconstruction, annotation, fitting, and rendering. Versioned records retain source hashes, code, fitted states, and evaluations underlying all tables and figures. Fitted quantities and selections are frozen before future scoring.

\section{Discussion and Limitations}
Our evaluation measures agreement with offline reconstructed tracks. Fitting and
scoring share depth/tracking errors, smoothing, and visibility-based selection,
including in image space. Independent geometric validation is unavailable,
and only 2 calibrated stereo clips support physical-unit errors.

Generality remains limited to 12 development-selected clips from 10 surgeries,
with periodic studies on 3 manipulation pilots and 1 passive recording. Pilot
outcomes informed the fitting procedure, so residual gains require fixed-procedure
testing on untouched clips against a similarly regularized nonperiodic residual
baseline.

The model assumes fixed grasp and camera, with jaw-translation input and effective
surface couplings. Slip, changing contact, and tool rotation require richer models.
Robust online use through occlusions also requires causal perception, observation
updates, and closed-loop evaluation.

\section{Conclusion}
Observation referencing provides the main improvement in our tool-conditioned tissue forecasts, with a smaller additional benefit from fitting and competitive simple baselines. Passive periodic forecasting improves at some cutoffs, while manipulation residual gains are clip-dependent and more limited under nested selection. The shared-cutoff protocol supports fixed-procedure evaluation on untouched clips with independent geometry, followed by online observations and prospective robotic tasks.

\bibliographystyle{ieeetr}
\bibliography{IEEEabrv, references}

\end{document}